\documentclass[a4paper]{article}
\usepackage{iwslt18,amssymb,amsmath,epsfig}
\def\reg{{\rm\ooalign{\hfil
     \raise.07ex\hbox{\scriptsize R}\hfil\crcr\mathhexbox20D}}}

\usepackage{times}
\usepackage{latexsym}
\usepackage{url}
\usepackage{xspace}
\usepackage{relsize}
\usepackage{float}
\usepackage{hyperref}
\usepackage{color}
\usepackage{graphicx}
\usepackage{subcaption}
\usepackage{booktabs}
\usepackage{multirow}
\usepackage{mathtools}
\usepackage{amsmath,amssymb}
\usepackage{latexsym}
\usepackage{cite}
\usepackage{adjustbox}
\usepackage{cleveref}

\usepackage{tikz,pgfplots}
\pgfplotsset{compat=newest}
\usetikzlibrary{plotmarks, shapes, arrows, positioning, shadows, trees}

\usepackage{mathptmx}
\usepackage{arabtex}
\usepackage{wrapfig}

\usepackage{fp}
\usepackage{hyperref}
\usepackage{paralist}
\usepackage{mdwlist}
\usepackage{CJKutf8}
\usepackage{footnote}

\newcommand\BLEU{\textsc{Bleu}\xspace}
\newcommand\TER{\textsc{Ter}\xspace}

\newcommand\WER{\textsc{Wer}\xspace}

\newcommand{\N}{\mathbb{N}}

\newlength\figureheight 
	\newlength\figurewidth

\makeatletter

\renewcommand{\subsection}{\@startsection
  {subsection}%
  {2}%
  {}%
  {-0.7\baselineskip}%
  {0.3\baselineskip}%
  {}}%

\renewcommand{\subsubsection}{\@startsection
  {subsubsection}%
  {3}%
  {}%
  {-0.7\baselineskip}%
  {0.3\baselineskip}%
  {}}%

\g@addto@macro\normalsize{%
  \setlength\abovedisplayskip{4pt plus 2pt minus 2pt}
  \setlength\belowdisplayskip{4pt plus 2pt minus 2pt}
  \setlength\abovedisplayshortskip{4pt plus 2pt minus 2pt}
  \setlength\belowdisplayshortskip{4pt plus 2pt minus 2pt}
}

\makeatother

\title{On Using SpecAugment for End-to-End Speech Translation}

 \makeatletter
 \def\name#1{\gdef\@name{#1\\}}
 \makeatother
\name{\em Parnia Bahar$^{1,2}$, Albert Zeyer$^{1,2}$, Ralf Schl{\"u}ter$^1$ and Hermann Ney$^{1,2}$}

\address{
$^1$Human Language Technology and Pattern Recognition Group \\
Computer Science Department, RWTH Aachen University, 52062 Aachen, Germany \\
$^2$AppTek, 52062 Aachen, Germany  \\ %\url{http://www.apptek.com/}\\
\texttt{\small \{bahar, zeyer, schlueter, ney\}@cs.rwth-aachen.de} \\
} 

\begin{document}
\maketitle
\begin{abstract}
This work investigates a simple data augmentation technique,
SpecAugment, for end-to-end speech translation.
SpecAugment is a low-cost implementation method applied directly to
the audio input features and it consists of masking blocks of frequency channels,
and/or time steps.
We apply SpecAugment on end-to-end speech translation tasks
and achieve up to +2.2\% \BLEU on LibriSpeech Audiobooks En$\to$Fr
and +1.2\% on IWSLT TED-talks En$\to$De
by alleviating overfitting to some extent.
We also examine the effectiveness of the method in a variety of data scenarios
and show that the method also leads to significant improvements in various data conditions
irrespective of the amount of training data.
\end{abstract}

\section{Introduction}
Traditional speech-to-text translation (ST) systems
have been build in a cascaded fashion comprised
of an automatic speech recognition (ASR) model
trained on paired speech-transcribed data
and a machine translation (MT) model
trained on bilingual text data.
Recent advancements
in both ASR \cite{bahdanau_2016_asr, chan_2016_las, chorowski_2015_attention_asr, zeyer_2018_att_asr, bahar_2019_2d_asr, zeyer2019:trafo-vs-lstm-asr}
and MT \cite{sutskever_14_seq2seq, bahdanau_15_attention, luong_15_attention, vaswani_17_transformer, bahar_2018_2d_mt}
have inspired the end-to-end direct ST models
which can be trained using a translation speech corpus \cite{weiss_2017_directly, berard_2016_proof}.
Some appealing advantages of the direct models are:
(1) no error accumulation from the recognizer,
(2) faster decoding throughput and
(3) less computational power in total by training all parameters jointly.
In spite of these properties,
training such end-to-end ST models requires a moderate amount
of paired translated speech-to-text data which is not easy to acquire.
Therefore these models tend to overfit easily.

In the absence of an adequate volume of training data,
one remedy is generating synthetic data like back-translation (BT) \cite{Sennrich16:monolingual} as
the most common data augmentation method to leverage monolingual data.
% in similar low-resource scenarios. 
The idea is to use a pre-trained model to convert weakly supervised data
into speech-to-translation pairs for ST training \cite{jia_2019_synthatic}.
One way is to use a pre-trained source-to-target MT model
to translate ASR transcription into the target language.
Another method is the use of a pre-trained text-to-speech (TTS) model
to generate speech data from a monolingual text.
However, these methods require some effort to train an additional model,
as well as computational power to generate a moderate amount of (noisy) synthetic data,
which in some cases can be too expensive to be obtained.

Another method is data augmentation
by which new synthetic training samples are generated by corrupting the initial audio data
and conserving the same label as the original training sample.
Audio-level speech augmentation can be done in different ways
such as noise injection (adding random noise),
shifting time (transmitting time series forward/backward with a few seconds),
speed perturbation (expanding time series by a speed rate)
and changing the frequency pitch randomly.
Besides increasing the quantity of training data,
data augmentation often make the model invariant to the applied noise and enhance its ability to generalize.

Inspired by the success of augmentation methods in ASR \cite{park2019specaugment, KoPPK15_dataaug},
as a remedy to avoid overfitting while using low-resource translated speech data,
we study the use of spectrogram augmentation (SpecAugment) for direct ST model.
SpecAugment \cite{park2019specaugment} is a simple and low-implementation cost approach.
Unlike traditional speech augmentation methods that directly manipulate the input signal,
SpecAugment is applied on the audio features,
which are usually mel spectrogram of the input signal.
We utilize two kinds of deformations of the spectrogram which are time and frequency masking,
where we mask a block of consecutive time steps and/or mel frequency channels.

Our main motivation of using SpecAugment is the potential avoidance of overfitting,
better generalization beyond low-resource training data and improving robustness of the end-to-end models. 
In this paper, we aim to shed light on the following
questions.
First, does SpecAugment strategy help the direct ST model?
Second, what is the effect of the approach concerning the different amount of training data?
Our first contribution is an extensive empirical investigation of SpecAugment
on top-performing ST systems to validate or disprove the above conjectures.
Our aim is not to compare with other data augmentation strategies,
but the effectiveness of the SpecAugment as a stand-alone method.
We hope that this method might overcome the data efficiency issue and therefore,
as our second contribution, we explore the effect of that on a various amount of training data.
Our experimental results on LibriSpeech Audiobooks En$\to$Fr and IWSLT TED-talks En$\to$De
show that the method not only greatly outperforms direct ST model up to +1.7\% \BLEU on average,
but also diminishes the overfitting problem.
We also show that our improvements are valid in different data scenarios irrespective of the amount of training data.

\section{Spectrogram Augmentation}
\label{sec:specaug}

% Since ST tasks are usually low-resource scenarios, we study a data augmentation technique similar to SpecAugment \cite{park2019specaugment} for direct end-to-end speech translation.
In spectrogram augmentation (SpecAugment) \cite{park2019specaugment},
we randomly apply masking in consecutive frames in the time axis as well as consecutive dimensions in the feature axis.
Since the author stated that the time warping is the most expensive and the least influential,
we do not explore it here. 

\begin{figure}
\centering
 \includegraphics[width=8cm,height=3cm,keepaspectratio]{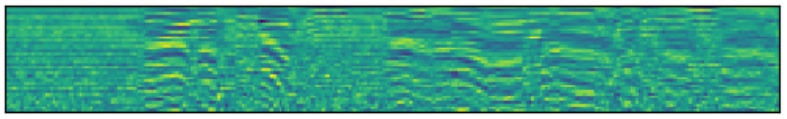}
 
  \includegraphics[width=8cm,height=3cm,keepaspectratio]{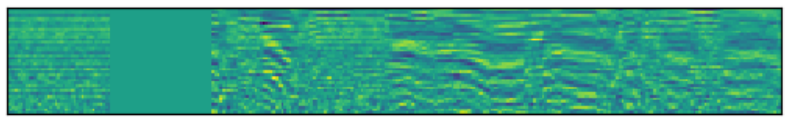}
  
  \includegraphics[width=8cm,height=3cm,keepaspectratio]{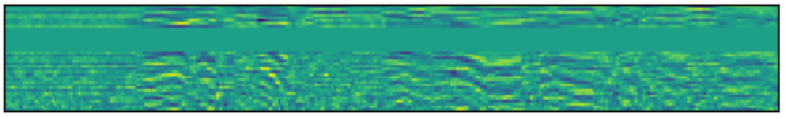}
  
  \includegraphics[width=8cm,height=3cm,keepaspectratio]{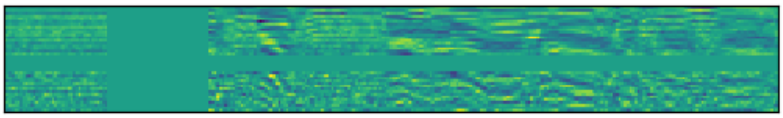}
  \caption{%
  From top to bottom, the figures depict the spectrogram
of the input with no augmentation, time masking, frequency
masking and both masking applied.}
\label{fig:specaug}
\end{figure}

\subsection{Time Masking}

Time masking is masking of $\tau$ successive time steps $[t, t + \tau)$,
where we set $(x_t, \dots, x_{t +\tau}) := 0$,
where $\tau$ is the masking window which 
is selected from a uniform distribution
from $0$ to the maximum time mask parameter $R$.
$(x_1,\dots,x_T)$ are the input audio features,
and $T$ is the length of the input signal.
The time position $t$ is picked from
another uniform distribution over $[0, T)$\footnote{We choose time position differently from the original paper where they select $t$ in the interval of $[0, T - \tau)$ \cite{park2019specaugment}.} such that we never exceed the maximum sequence length $T$ (i.e. if $t +\tau > T$, we set it to $T$).

We apply the time masking procedure for $m_R \in \N_0$ times. 
We also ensure that if $m_R > 1$, the same time position $t$ is not selected more than once (i.e. without replacement)\footnote{It is not clear whether the original paper allows replacement or not.}.

\subsection{Frequency Masking}

Frequency masking can be also applied such that $\phi$ consecutive frequency
channels $[f, f +\phi)$ are masked,
where $\phi$ is chosen from
a uniform distribution from 0 to the frequency mask parameter
$F$, and $f$ is chosen from $[0, \nu)$\footnote{Again we note the difference between our implementation and the original paper where the selection interval is $[0, \nu-\phi)$ \cite{park2019specaugment}.}.
$\nu$ is the input feature dimension,
e.g.~the number of mel frequency channels. Similar to time masking, we do not allow for already selected $f$ and check if $f +\phi > \nu$, we set it to $\nu$.

Figure \ref{fig:specaug} shows examples of the individual augmentations applied
to a single input. 
Multiple frequency and time masks might overlap.
$m_F \in \N_0$ refers to the number of times we apply the frequency masks.
We note that we standardize the features to zero mean and variance of one.
Therefore, masking to zero is equivalent to setting it to the mean value.
In this work, we mainly investigate a series of combinations to find a reliable recipe for direct ST model.
We only apply the SpecAugment during training.

\section{Network}
\label{sec:network}

\begin{figure}
  \centerline{\includegraphics[width=3.2cm,height=6.5cm,keepaspectratio]{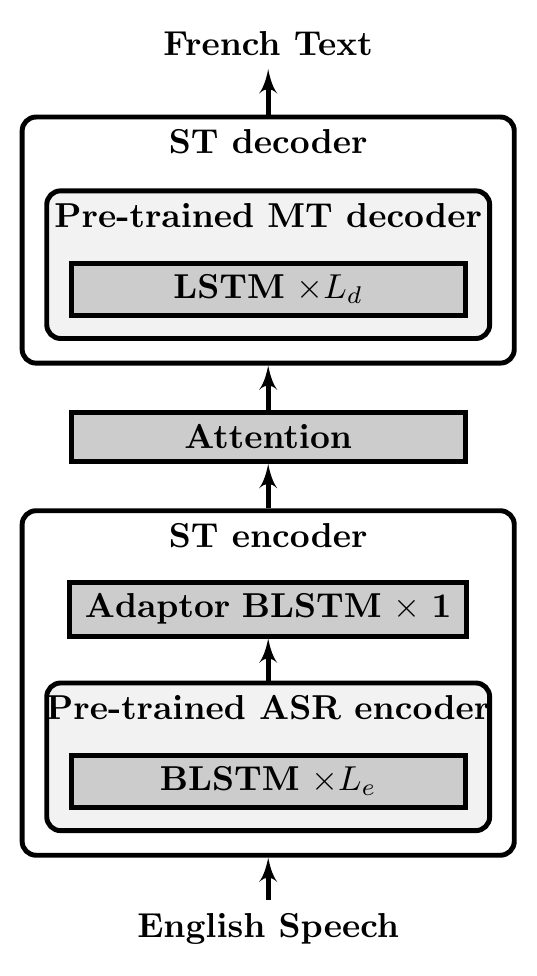}}
\caption{%
Overview of the direct speech translation model.
Shallow grey
blocks correspond to pre-trained components, 
and dark grey blocks are fine-tuned on the ST task.}
 \label{fig:direct}
\end{figure}

In ST, given an input observation (audio feature vectors) of variable length $T$, $x_1^{T}$,
a sequence of discrete label of unknown length $J$ (source sequence/transcribed words),
$f_1^{J}$ and a sequence of target words of unknown length $I$,
$e_1^{I}$, the posterior probability of a target sequence is defined as:
\begin{align}
p(e_1^{I} | x_1^{T}) = \prod_{i=1}^{I} p(e_i| e_1^{i-1}, x_1^{T})
\end{align}
where usually $T > I$, $J$.
This posterior can be modeled directly in an end-to-end fashion.
Here, we only address the direct end-to-end architecture that is used in our experiments.

The direct model is based on the attention sequence-to-sequence model \cite{bahdanau_15_attention}
composed of long short-term memories (LSTMs) \cite{hochreiter_97_LSTM} similar to \cite{bahar2019:st-comparison}.  
We only focus on LSTM-based models and leave the transformer architecture 
as our future study \cite{vaswani_17_transformer, gangi-etal-2019-enhancing,st_kd_interspeech2019}.
An abstract overview of the network and a summary of the model are shown in Figure \ref{fig:direct} 
and written in Equation \ref{eq:att} respectively.
A bidirectional LSTM (BLSTM) scans the input sequence once from left to right and once from right to left.
To handle the long speech utterances,
we apply max-pooling in the time-dimension at multiple steps inside the speech encoder. 
% Similar to \cite{zeyer_2018_att_asr}, we apply layer-wise pre-training for the speech encoder, where we start with two encoder layers and a single max-pool in between BLSTM layers. 
For the input sequence $x_1^T$,
we end up with the encoder states $h_1^{T'}$,
where $T' = T / {red}$ with the time reduction factor ${red}$.
%We apply 2 max-pooling layers with pool-size of 3 and 2,
%i.e.~we get a total time reduction factor of $\rm{red} = 6$.
Then, the decoder LSTM generates an output sequence conditioned on the encoder representations.
While computing $e_{i}$ at each time step,
an additive attention function is used to produce normalized attention weights $\alpha_{i,t}$.
% which can be viewed as alignment probabilities that score how likely the input frame, $x_n$, is aligned to the current output token, $e_{i}$. 
The context vector $c_i$ is then computed as a weighted sum of encoder representations.
A transformation followed by a softmax operation predicts $e_{i}$.
Finally, the decoder state is updated to $s_i$.
Here, $L_e$ and $L_d$ are the number of encoder and decoder layers respectively.
\(\circ\) is the concatenation operator of functions.
\begin{align}
&h_1^{T'} = (\operatorname{BLSTM}_{\rm{L_e}}
\circ \cdots
\circ \operatorname{max-pool}_1
\circ \operatorname{BLSTM}_1) (x_1^T) \nonumber \\
&\alpha_{i,t} = \operatorname{softmax}_{\overline{t}}
    \big( \operatorname{linear} ( \tanh ( \operatorname{linear}( s_{i-1},h_{t} ))) \big) \nonumber \\
&c_i = \sum_{t=1}^{T'} \alpha_{i,t} h_t  \nonumber \\
&p(e_{i}|e_1^{i-1},x_1^T) = \operatorname{softmax}_{\overline{e}}
    \big( \operatorname{linear}(e_{i-1}, s_{i-1}, c_{i}) \big)  \nonumber\\
&s_i = \operatorname{LSTM}_{L_d}
\circ \cdots
\circ \operatorname{LSTM}_1 (e_{i}, s_{i-1}, c_{i}) \label{eq:att}
\end{align}

\section{Experiments}
\label{sec:expriments}

\subsection{Datasets and Metrics} 
\label{sec:dataset}

We have conducted our experiments on two ST tasks which are publicly available: the LibriSpeech Audio-books En$\to$Fr \cite{Kocabiyikoglu_2018_librispeech, berard_2018_librispeech}\footnote{https://persyval-platform.univ-grenoble-alpes.fr/DS91/detaildataset} and the IWSLT TED-talks En$\to$De \cite{Cho_2014_ted,Niehues_2018_iwslt2018}\footnote{https://sites.google.com/site/iwsltevaluation2018/Lectures-task}. The training data statistics are listed in Table \ref{tab:stat}.

\begin{table}[h]
\begin{center}
\caption{Training data statistics.}
\scalebox{0.9}{%
\label{tab:stat}
\begin{tabular}{lrr|rr}
\hline
\multirow{2}{*}{\bfseries Task}  & \multicolumn{2}{c}{\bfseries LibriSpeech En$\to$Fr} & \multicolumn{2}{c}{\bfseries IWSLT En$\to$De}\\ \cline{2-5}
     & \# of seg.  & hours    & \# of seg.    & hours    \\ \hline
ASR  &  61.3k  &   130h     &  92.9k    &  207h    \\
ST   &  94.5k  &   200h &  171.1k   &  272h        \\
MT   &  94.5k   &  - &  32M    &   -    \\ \hline        
      
\end{tabular}
}
\end{center}
\end{table}

\textbf{LibriSpeech En$\to$Fr:}
% As it is seen in Table \ref{tab:stat}, the ASR and ST data for the LibriSpeech is smaller than that of the IWSLT. 
As suggested by \cite{berard_2018_librispeech}, 
to double the training data size,
we concatenate the original translation and the Google Translate reference
which have been provided in the dataset package.
Hence, we end up to 200h of clean speech corresponding to 94.5k segments for the ST task.
We apply 40-dimensional Gammatone features \cite{schluter_2007_gammatone} using
the RASR feature extractor \cite{wiesler_2014_rasr}.
For MT training, we utilize no additional data and only use the source-target data from the ST task, i.e. 94.5k. 
For ASR training, we take both the ASR and ST data resulting in 330h.
The dev and test sets contain 2h and 4h of speech, 1071 and 2048 segments respectively.
Here, the dev set is used as our cross-validation set as well as checkpoint selection.

\textbf{IWSLT En$\to$De:}
Similar to \cite{bahar2019:st-comparison, apptek_2018_st}, we extract 80-dimensional Mel-frequency cepstral coefficients (MFCC) features. 
We automatically recompute the provided audio-to-source-sentence alignments
to reduce the problem of speech segments without a translation.
We use the TED-LIUM corpus including 207h and the IWSLT speech translation TED corpus
with 272h of speech data for ASR training. 
For MT training,
we use the TED, and the OpenSubtitles2018 corpora,
as well as the data provided by the WMT 2018 evaluation
(Europarl, ParaCrawl, CommonCrawl, News Commentary, and Rapid),
a total of 65M lines of parallel sentences.
We filter these data based on several heuristics resulting in 32M samples.
We randomly select a part of our segments as our cross-validation set
and choose dev2010 and test2015 as our dev and test sets with 888 and 1080 segments respectively.
We select our checkpoints based on dev2010 set.

For both tasks, we remove the punctuation from the transcriptions (i.e. the English text)
and keep the punctuation on the target side. 
After tokenization using
\texttt{Moses} toolkit \cite{koehn_07_moses}\footnote{http://www.statmt.org/moses/?n=Moses.SupportTools},
we apply frequent casing for the IWSLT tasks while lowercase for the LibriSpeech data.
Therefore, the evaluation of the IWSLT En$\to$De is case-sensitive,
while that of the LibriSpeech is case-insensitive\footnote{%
We do the case-insensitive evaluation to be comparable with the other works,
however, it is not clear which \BLEU script they used \cite{chung_2019_unsupervised_st, berard_2018_librispeech}.}.
The translation models are evaluated using the official scripts of WMT campaign,
i.e. \BLEU~\cite{papineni_02_bleu} computed by
\texttt{mteval-v13a}\footnote{ftp://jaguar.ncsl.nist.gov/mt/resources/mteval-v13a.pl}
and \TER~\cite{snover_06_ter} computed by \texttt{tercom}\footnote{http://www.cs.umd.edu/~snover/tercom/}.
\WER~is computed by \texttt{sclite}\footnote{http://www1.icsi.berkeley.edu/Speech/docs/sctk-1.2/sclite.htm}.

\subsection{Models} 
\label{sec:models}

In our experiments, we build ASR, MT, and ST models all based on the network described in \Cref{sec:network}.
The ASR and MT models are used for building the cascade pipeline as well as pre-training the components of the ST model.
Thus, the ASR and ST models use the same speech encoder architecture,
whilst the MT and ST models use the same text decoder topology, as illustrated in Figure \ref{fig:direct}.

For both tasks, we apply separate byte pair encoding (BPE) \cite{sennrich_16_bpe}
with $20$k symbols on both side of the MT data,
whereas $10$k merge operations on the ASR transcriptions. \\
\textbf{ASR model:} All tokens are mapped into a $512$-dimensional embedding space.
The encoder is composed of 6 stacked BLSTM layers with 1024 nodes.
The decoder is a 1-layer unidirectional LSTM of size 1024.
A single head additive attention with alignment feedback \cite{tu2016ACL,bahar_2017_rwth}
is used as our attention component.
Similar to \cite{bahar2019:st-comparison, zeyer_2018_att_asr},
we apply layer-wise pre-training for the encoder,
where we start with two encoder layers and a single max-pool in between BLSTM layers. 
We apply 2 max-pooling layers with pool-size of 3 and 2 , i.e.~we get a total time reduction factor of 6. 
We also use CTC auxiliary loss function \cite{graves_2006_ctc} on top of the speech encoder only during training \cite{hori2017attctc}.
\\
\textbf{MT model:} Our MT model follows the ASR model with a 6-layer BLSTM encoder without max-pooling,
with a cell size of 1024. The decoder is a 1-layer unidirectional LSTM with cell size 1024,
with single head additive attention equipped with alignment feedback. \\
\textbf{ST model:} The encoder has a similar architecture to the ASR encoder,
and the decoder is similar to the MT decoder. 

The models are trained end-to-end using the Adam optimizer \cite{kingma_14_adam}
with a learning rate of 0.0008,
and a dropout of 0.3 \cite{srivastavad_14_dropout}. We warm-up the learning rate by linearly increasing it for a few training steps. 
Label smoothing \cite{pereyra_2017_label_smoothing} with a ratio of 0.1 is utilized.
We employ a learning rate scheduling scheme,
where we lower the learning rate with a decay factor of 0.9
if the perplexity on the dev set does not improve for 5 consecutive checkpoints and save the checkpoints every fifth of an epoch.
We remove sequences longer than 75 tokens before batching them together.
All batch sizes are specified to be as big as possible to fit in a single GPU memory. 
A beam size of 12 is used in inference.
In order to explore the impact of SpecAugment,
we choose to have relatively large models, as explained above.
These models include 181M and 192M free parameters for Librispeech and IWSLT tasks respectively.  
The models are built using our in-house NN-toolkit software that relies on TensorFlow \cite{tensorflow}.
The code is open source and the configurations of the setups are available online.

\begin{table}
\begin{center}
\caption{ASR results measured in \WER~[\%].}
\label{tab:asr_results}
\begin{tabular}{lrr}

\toprule
\multirow{2}{*}{\bfseries Task} & \multicolumn{2}{c}{\bfseries \WER[$\downarrow$]}  \\ \cline{2-3}
& \bfseries  dev  &  \bfseries test \\ \midrule
LibriSpeech En$\to$Fr &  \phantom{0} 6.47 & \phantom{0} 6.47  \\
IWSLT En$\to$De       &  12.36  &  13.80   \\ 
\bottomrule
\end{tabular}
\end{center}
\end{table}

\begin{table}
\begin{center}
\caption{MT results measured in \BLEU~[\%] and \TER~[\%] trained using ground truth source text.}
\label{tab:mt_results}
\begin{tabular}{lllll}
\toprule
\multirow{2}{*}{\bfseries Task} & \multicolumn{2}{c}{\bfseries \BLEU[$\uparrow$]} & \multicolumn{2}{c}{\bfseries \TER[$\downarrow$]}  \\ \cline{2-5}
& \bfseries  dev  & \bfseries  test &  \bfseries dev  &  \bfseries test\\ 
\midrule
LibriSpeech En$\to$Fr &  20.1 & 18.2 &  65.3  & 67.7 \\ 
IWSLT En$\to$De       &  30.5 & 31.5 &  50.6  & -  \\ 
\bottomrule
\end{tabular}
\end{center}
\end{table}

\begin{table}
\begin{center}
\caption{ST results using cascaded pipeline of independent ASR and MT models measured in \BLEU~[\%] and \TER~[\%]. We highlight that the cascaded pipeline used more training data compared to the direct model.}
\label{tab:st_results}
\begin{tabular}{lllll}
\toprule
\multirow{2}{*}{\bfseries Task} & \multicolumn{2}{c}{\bfseries \BLEU[$\uparrow$]} & \multicolumn{2}{c}{\bfseries \TER[$\downarrow$]}  \\ \cline{2-5}
& \bfseries  dev  & \bfseries  test &  \bfseries dev  &  \bfseries test\\ 
\midrule
LibriSpeech En$\to$Fr &  17.3 & 15.7 & 69.1   & 70.6 \\ 
IWSLT En$\to$De       &  24.7 & 24.4 & 58.9  &  62.5 \\ 
\bottomrule
\end{tabular}
\end{center}
\end{table}

\section{Results}
Table \ref{tab:asr_results} and \ref{tab:mt_results} present the results for the ASR and MT models
(described in Section \ref{sec:models}) on LibriSpeech and IWSLT tasks respectively.
On the test sets, we achieve 6.47\% and 13.80\% \WER.
We note that one might gain better \WER
using the conventional hybrid hidden Markov model (HMM) - neural network (NN) approach
on phoneme level \cite{bourlard1994hybrid},
which is out of the scope of this paper. 

The MT task on LibriSpeech seems more challenging as both scores are lower.
We obtain 18.2\% \BLEU on the LibriSpeech and respectively 31.5\% \BLEU on the IWSLT by pure MT.
Table \ref{tab:st_results} shows the traditional cascade pipeline where the output of our ASR model,
a sequence of tokens is fed as the input to our MT system.
We gain 15.7\% \BLEU and 70.6\% \TER on the LibriSpeech and 24.4\% \BLEU and 62.5\% \TER on IWSLT test set.
As expected, the ST systems are behind the pure MT models (cf.~\Cref{tab:mt_results} and \Cref{tab:st_results}).
In the rest of the paper, we only focus on the results of direct models.

\subsection{Using SpecAugment}

In this section, we explore different types of masking.
In the first set of experiments, we deactivate either time or frequency masking. 
The results are listed in Table \ref{tab:specAug_libri} and \ref{tab:specAug_iwslt} for LibriSpeech and IWSLT tasks respectively.
As it is shown in Table \ref{tab:specAug_libri}, we apply various types of frequency masking with different values of $F$, between 2 and 35,
and $m_F$ while the time masking is disabled. 
% $m_F$ represents the number of times this masking has been applied. 
As listed,
the optimum value of $F$ is around 4 and 5 with 1.5\% in \BLEU and 1.4\% in \TER on average of dev and test sets.
A further increase of $F$ until 20 hurts the performance by 0.4\% in both \BLEU and \TER on the test set.
To verify the effect of SpecAugment, we also employ a coarse policy where 
we randomly mask 35 frequency channels out of 40 and apply it 5 times.
As expected, the performance drops behind the direct baseline.
Interestingly enough, even a small value of frequency masking ($F=2$) leads to improvements.
It is important to highlight that since we randomly select the masking window between zero and maximum value of $F$,
the results are close to each other. The best results of each set of experiments are highlighted in bold.

\begin{table}
\begin{center}

\caption{SpecAugment results on LibriSpeech En$\to$Fr using various augmentation parameters.
\lq\lq-\rq\rq: failed training. The bold augmentation parameters are used in the rest of our experiments.}
\adjustbox{max width=0.5\textwidth}{
\vspace{0.2cm}
\begin{tabular}{rrrrrrrr}

\toprule
\multirow{2}{*}{\bfseries $F$} & \multirow{2}{*}{\bfseries $m_F$} & \multirow{2}{*}{\bfseries $R$} & \multirow{2}{*}{\bfseries $m_R$} & \multicolumn{2}{c}{ {\bfseries \BLEU} [\%]} & \multicolumn{2}{c}{ {\bfseries \TER} [\%]} \\ \cline{5-8}
&&&& \bfseries dev & \bfseries test & \bfseries dev & \bfseries test \\ \hline

0 & 0 & 0 & 0 & 15.8 & 15.2 & 74.1& 75.8\\

2  & 1 & 0 & 0 & 17.3 & \textbf{16.1} &  72.5 &  75.0\\

4  & 1 & 0 & 0 & \textbf{18.0} & \textbf{16.1} &  \textbf{72.2} &  74.9\\

4  & 2 & 0 & 0 & 17.6 & \textbf{16.1} &  72.3 & \textbf{74.6} \\
4  & 4 & 0 & 0 & 17.1 & \textbf{16.1} & 73.2  & \textbf{74.6} \\
5  & 1 & 0 & 0 & 17.6&  \textbf{16.1}& 72.4 & 74.7 \\
8   & 1 & 0 & 0 & 17.4&  15.7& 72.8 & 75.2\\
8   & 2 & 0 & 0 & 17.5&  15.7& 73.0 & 75.9\\
20  & 1 & 0 & 0 & 17.2&  15.7 & \textbf{72.2} & 75.0  \\
20  & 2 & 0 & 0 & 16.7&  15.4 & 73.0 & 75.9   \\
35  & 5 & 0 & 0 & 16.1 & 15.0 & 74.1 & 76.8 \\

\hline
0 & 0 & 20 & 1 & 17.2  & 15.7 & 73.3 &  75.5 \\
0 & 0 & 20 & 2 & 17.3  & \textbf{16.2} & 72.9 &  75.2 \\

0 & 0 & 40 & 1 &  17.4 & 15.6 & 73.2 & 75.8  \\
0 & 0 & 40 & 2 &  17.3 & \textbf{16.2} & \textbf{72.5} & \textbf{74.7}  \\ 

0 & 0 & 40 & 4 &  16.6 & 15.6 & 73.1& 75.9  \\ 
% 0 & 0 & 60 & 1 &  16.6 & 15.8 & 74.0 & 75.7  \\
% 0 & 0 & 60 & 2 &  17.4 & 15.8 & 73.1 & 75.4 \\
0 & 0 & 80 & 1 &  \textbf{17.5} & \textbf{16.1} & \textbf{72.5} & 75.1  \\

0 & 0 & 100 & 1 & 17.3 & 15.9 & 73.8 & 75.7 \\
0 & 0 & 100 & 2 & 16.8 & 15.4 & 74.1 & 76.6 \\

0 & 0 & 200 & 2 & 16.6 & 15.3 & 73.8 & 76.0 \\
0 & 0 & 400 & 5 & - & - & - & - \\

\hline
% 5 & 1 & 60 & 2 &  17.4 & 15.8 & 72.6 & 74.9  \\
\textbf{5} & \textbf{1} & \textbf{40} & \textbf{2} &  \textbf{17.9} & \textbf{16.1} & \textbf{72.2} & \textbf{74.6}  \\
5 & 1 & 80 & 1 &  17.6 & 15.6 & 72.6 & 75.5  \\

4 & 1 & 40 & 2 & \textbf{17.8}  & \textbf{16.1}  & 72.6  & 75.1  \\
4 & 2 & 40 & 2 & 16.8  & 15.7  &  72.3 & 74.6  \\

\bottomrule
\end{tabular}
}
\label{tab:specAug_libri}
\end{center}

\end{table}

We also disable the frequency masking by setting $F$ and $m_F$ set to zero
and vary the time mask parameter $R$ and the number of times it has been called.
Again, 
there is a limit of how much data augmentation can be applied.
Enlarging the time masking window $R$ to 100 leads to lower \BLEU and \TER scores.
Furthermore, we drastically increase the time masking window $R$ into 400 steps
and apply it 5 times which fails due to unstable training.
If the initial convergence is stable,
in all cases, adding some data augmentation improves the setup,
however, at some point, the performance degrades by more augmentation. 

Moreover, we observe that in many cases applying small window several times
gives slightly more improvements compared to the policy in which a large window applied once
(cf.~row 4 and 7 in Table \ref{tab:specAug_libri}).
After finding the optimum of both time and frequency masking,
we have done some combinations of both masking as shown in the table.  
As expected, the combination of best of two masking gives the largest boost.

\begin{table}
\begin{center}

\caption{SpecAugment results on IWSLT En$\to$De using various augmentation parameters.}
\adjustbox{max width=0.5\textwidth}{
\vspace{0.2cm}
\begin{tabular}{rrrrrrrr}

\toprule
\multirow{2}{*}{\bfseries $F$} & \multirow{2}{*}{\bfseries $m_F$} & \multirow{2}{*}{\bfseries $R$} & \multirow{2}{*}{\bfseries $m_R$} & \multicolumn{2}{c}{ {\bfseries \BLEU} [\%]} & \multicolumn{2}{c}{ {\bfseries \TER} [\%]} \\ \cline{5-8}
&&&& \bfseries dev & \bfseries test & \bfseries dev & \bfseries test \\ \hline
0 & 0 & 0 & 0 & 16.9 & 16.5 & 67.3& 70.6\\

4  & 1 & 0 & 0 & 17.8 & 17.0  & 66.3 & 69.5 \\
4  & 2 & 0 & 0 & 17.3 & 17.4 & 66.5& 69.3 \\

5  & 1 & 0 & 0 & \textbf{18.1} & \textbf{17.5} & 66.3 & \textbf{68.8} \\

8  & 1 & 0 & 0 & 17.6 & 17.1 & \textbf{66.2} & - \\
10  & 1 & 0 & 0 & 16.9 & 16.9 & 66.9 & 70.2  \\
20  & 1 & 0 & 0 & 17.0 & 17.1 & - & 70.3 \\
% 40  & 1 & 0 & 0 & 17.7 & 17.1 & 65.9 & 69.7 \\

\hline
0 & 0 & 20 & 1 & 17.7 & \textbf{17.6} & - &  70.1 \\
0 & 0 & 20 & 2 & 17.5  & 17.5  & \textbf{65.8}  & \textbf{69.2}  \\

0 & 0 & 40 & 1 & 17.5 & 17.1 & 66.1 & 70.0 \\
0 & 0 & 40 & 2 & \textbf{17.9} & 17.5 & \textbf{65.8} & 69.5 \\

0 & 0 & 60 & 1 & 17.6 & 17.3 & 66.6 &  - \\

0 & 0 & 80 & 1 & 16.9 & 17.1 & 68.6 & 69.9 \\
0 & 0 & 80 & 2 & 16.8 & 16.6 & 68.0 & 72.0 \\

0 & 0 & 100 & 1 & 17.7 & 17.0 & 66.2 & -\\

\hline
5 & 1 & 40 & 2 & 17.8 & 16.8 & 66.2 & 70.9 \\
% 5 & 1 & 20 & 1 &  17.6 & 16.9  & 66.0  & 69.4 \\

4 & 1 & 40 & 1 & 17.5  &  17.2 &  66.5 &  69.4\\
\textbf{4} & \textbf{1} & \textbf{40} & \textbf{2} & 17.7  &  \textbf{18.0}  & \textbf{66.0} & \textbf{69.2} \\

\midrule

27 & 2 & 100 & 2 & 16.6 & 16.7 & 67.6& 70.4\\

\bottomrule

\end{tabular}
}
\label{tab:specAug_iwslt}
\end{center}
\end{table}

We verify the influence of SpecAugment on IWSLT En$\to$De task
by an improvement up to 1.2\% in \BLEU and 1.8\% in \TER (see Table \ref{tab:specAug_iwslt}).
For IWSLT, we also apply a policy
similar to the \texttt{LD} policy of main paper \cite{park2019specaugment}
as listed in the last row of Table \ref{tab:specAug_iwslt}.
As shown, this setup is not the optimum case for our task,
which leads to the conclusion
that SpecAugment might be working better by fine-tuning on a specific task,
however, adding some data augmentation improves the setup.
Moreover, based on the above results,
SpecAugment performs quite well regardless of the features and their dimensions.
In our experiments, we use 40-dimensional Gammatone
features for LibriSpeech respectively 80-dimensional MFCC
features for IWSLT. 
In both cases, augmentation helps the performance.
In the rest of our experiments, we use the augmentation parameters which is bold in the tables. 
For IWSLT, we choose $F=4$ rather $F=5$, but the rest of the parameters are the same.

\begin{figure}[h]
\centering
	\setlength\figureheight{3cm} 
	\setlength\figurewidth{1.0\textwidth}
	    % GRAPHICS
\pgfplotsset{width=6cm, height=4.0cm,compat=1.3}
\definecolor{green}{RGB}{100,200,50}
% \definecolor{applegreen}{rgb}{0.0, 0.26, 0.0}

\begin{tikzpicture}

\begin{axis}[
  scale only axis,
  xmin=0, xmax=21,
  ymin=0.5, ymax=5,
  xlabel=Epochs,
  ylabel=$\log$ PPL,
  legend style={legend cell align=left,align=left,draw=white!15!black, font=\scriptsize, legend pos=north east, style={row sep=0.005cm}},]
]

\addplot[dashed,line width=1.2pt, mark size=1pt, blue] 
  coordinates{
( 0 , 5.852225684082929 )
( 1 , 4.4168490121493775 )
( 2 , 3.928273357889215 )
( 3 , 3.6879911938255514 )
( 4 , 3.3133074415086354 )
( 5 , 2.760172079509073 )
( 6 , 2.6737742250700043 )
( 7 , 2.6344178338052173 )
( 8 , 2.565136697811174 )
( 9 , 2.5819616916047163 )
( 10 , 2.5676531255075097 )
( 11 , 2.570129676164184 )
( 12 , 2.6100830754389164 )
( 13 , 2.6141574165442036 )
( 14 , 2.6399685212945307 )
( 15 , 2.65660533052639 )
( 16 , 2.6770924779758136 )
( 17 , 2.7037531079667194 )
( 18 , 2.7209252984942354 )
( 19 , 2.7450967622891485 )
}; \label{plot_one}
\addlegendentry{dev - w SpecAug};

% \addlegendimage{/pgfplots/refstyle=plot_one}\addlegendentry{dev - w SpectAug}

\addplot[dashed,line width=1.2pt, mark size=1pt, green]
  coordinates{
( 0 , 5.7240089317210145 )
( 1 , 4.746090170535346 )
( 2 , 3.204445560227776 )
( 3 , 2.8563606705130096 )
( 4 , 2.730920074958011 )
( 5 , 2.68329316130448 )
( 6 , 2.650961985869247 )
( 7 , 2.6335362273599188 )
( 8 , 2.6437582193940803 )
( 9 , 2.645482725559778 )
( 10 , 2.6806982236524854 )
( 11 , 2.718851057572259 )
( 12 , 2.749106571308294 )
( 13 , 2.78661919102564 )
( 14 , 2.8218783621996755 )
( 15 , 2.848041752046702 )
( 16 , 2.8973094426311534 )
( 17 , 2.9064221328373407 )
( 18 , 2.933947727057543 )
( 19 , 2.954480020560122 )
}; \label{plot_two}
% \addlegendimage{/pgfplots/refstyle=plot_two}\addlegendentry{dev - w/o SpectAug}
\addlegendentry{dev - w/o SpecAug};

\addplot[solid,line width=1.2pt, mark size=1pt, blue]
  coordinates{
( 0 , 6.337640411345245 )
( 1 , 4.532284852494809 )
( 2 , 4.008715375216289 )
( 3 , 3.9286610098043315 )
( 4 , 3.1326636998512547 )
( 5 , 2.4906060029896553 )
( 6 , 2.2458767169915688 )
( 7 , 2.0679242241834035 )
( 8 , 1.860784562180753 )
( 9 , 1.7202693938649032 )
( 10 , 1.5780532963979301 )
( 11 , 1.4424169718710527 )
( 12 , 1.3452190809325961 )
( 13 , 1.232775695242522 )
( 14 , 1.171065121294334 )
( 15 , 1.0749294905109272 )
( 16 , 1.0058804754235666 )
( 17 , 0.9518641465455666 )
( 18 , 0.9006155377502818 )
( 19 , 0.8651569491422152 )
}; \label{plot_three}
% \addlegendimage{/pgfplots/refstyle=plot_three}\addlegendentry{train - w SpectAug}
\addlegendentry{train - w SpecAug};

\addplot[solid,line width=1.2pt, mark size=1pt, green]
  coordinates{
( 0 , 6.227096643809515 )
( 1 , 4.930567489060985 )
( 2 , 3.2010651922943802 )
( 3 , 2.6123447046853103 )
( 4 , 2.3061235160666493 )
( 5 , 2.091545748121485 )
( 6 , 1.881119348596891 )
( 7 , 1.699341644661724 )
( 8 , 1.5233873807609568 )
( 9 , 1.3535350192254778 )
( 10 , 1.2186866279601454 )
( 11 , 1.0960862963506148 )
( 12 , 0.9936151144483043 )
( 13 , 0.9157089255625217 )
( 14 , 0.8387929713547139 )
( 15 , 0.7823186849252329 )
( 16 , 0.7825241434272219 )
( 17 , 0.7214122809634366 )
( 18 , 0.6824709756883963 )
( 19 , 0.6529931274014021 )
}; \label{plot_four}
% \addlegendimage{/pgfplots/refstyle=plot_four}\addlegendentry{train - w/o SpectAug}
\addlegendentry{train - w/o SpecAug};

\end{axis}
\end{tikzpicture}
\caption{Average log perplexity of training and dev sets across epochs.}
\label{fig:overfitting}
\end{figure}
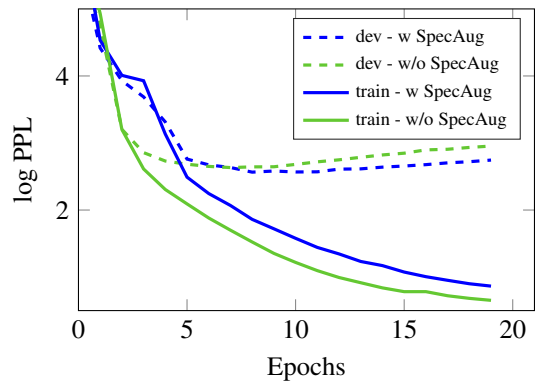

\begin{table}
\begin{center}
\caption{SpecAugment results with with a varying amount of training data on LibriSpeech En$\to$Fr.
$\Delta$ indicates the absolute difference on average of dev and test sets.}
\adjustbox{max width=0.45\textwidth}{
\vspace{0.2cm}
\begin{tabular}{lcccccc}
\toprule
\multirow{2}{*}{\bfseries \# segments} & \multicolumn{3}{c}{ {\bfseries \BLEU} [\%]} & \multicolumn{3}{c}{ {\bfseries \TER} [\%]} \\ \cline{2-7}
& \bfseries dev & \bfseries test & \bfseries $\Delta$ & \bfseries dev & \bfseries test & \bfseries $\Delta$ \\ 
\midrule
23k &  11.6 & 7.0 & \multirow{2}{*}{1.0} &  90.2& 92.3 &  \multirow{2}{*}{-1.2} \\
 \quad + SpecAug & 13.3 & 7.3 &  &  91.5 & 93.4  &   \\
 \midrule

47k &  14.8 & 10.2 & \multirow{2}{*}{1.8} & 87.2 &  89.1 & \multirow{2}{*}{2.8}\\
\quad + SpecAug & 17.3 & 11.2 &  & 83.4 &  87.4 &  \\
\midrule

71k & 15.4  & 13.2 & \multirow{2}{*}{1.1} & 78.9 &  81.0 & \multirow{2}{*}{2.1}\\
\quad + SpecAug & 16.7 & 14.4 &  & 76.7 & 79.1  &  \\

\midrule

94k &15.8 & 15.2 & \multirow{2}{*}{1.5} & 74.1& 75.8 & \multirow{2}{*}{1.6} \\
\quad + SpecAug & 17.9 & 16.1 & & 72.2 & 74.6 &  \\

\bottomrule
\end{tabular}
}
\label{tab:diff_training_data}
\end{center}
\end{table}

\begin{table*}
\begin{center}

\caption{SpecAugment results with pre-training,
which makes use of more training data.}
\adjustbox{max width=1.0\textwidth}{
\begin{tabular}{lcccccccc}
\toprule
\multirow{3}{*}{\bfseries method} & \multicolumn{4}{c}{\bfseries LibriSpeech En$\to$Fr} & \multicolumn{4}{c}{\bfseries IWSLT En$\to$De} \\
 & \multicolumn{2}{c}{ {\bfseries \BLEU} [\%]} & \multicolumn{2}{c}{ {\bfseries \TER} [\%]} & \multicolumn{2}{c}{ {\bfseries \BLEU} [\%]} & \multicolumn{2}{c}{ {\bfseries \TER} [\%]}\\ \cline{2-9}
& \bfseries dev & \bfseries test & \bfseries dev & \bfseries test & \bfseries dev & \bfseries test & \bfseries dev & \bfseries test \\ 
\midrule
direct & 15.8 & 15.2 & 74.1 & 75.8 & 16.9 & 16.5 & 67.3& 70.6 \\
\quad + pretraining & 18.0 & 15.8 & 71.3 & 73.9 & 21.1 & 20.7 & 62.1 &  65.5  \\
\quad\quad + SpecAugment & 18.5 & 16.2 & 71.0 &  74.5 & 21.3 & 20.9 & 61.9&  65.7 \\

\bottomrule
\end{tabular}
}
\label{tab:specAug_pretraining}
\end{center}
\end{table*}

\subsection{Importance of SpecAugment on overfitting}

We also study the effect of SpecAugment on overfitting.
Figure \ref{fig:overfitting} shows log-perplexity plots on training and dev
set with and without augmentation.
It is seen that SpecAugment leads to better generalization,
as measured from the difference between the perplexity of training and dev data.
The model trained with SpecAugment still has a training data likelihood which is higher than the baseline system.
Therefore, we can confirm that the method reduces overfitting up to some degree.
In this case, we need to train the SpecAugment system for few more epochs.
A corresponding increase in the number of epochs for the baseline system deteriorated the performance.
%Here, we split our epoch into 5 sub-epochs.

\subsection{Effect on training data size}

We go further and show that SpecAugment can be leveraged to improve the performance of a direct ST model,
when the amount of training data is limited.
To do so, we have conducted studies on a different portion of training data to see how the method performs
in different data conditions.
Table \ref{tab:diff_training_data} compares the performance improvement of SpecAugment
with a varying amount of training data.
It can be seen that it is helpful in all scenarios irrespective of the amount of training data. 

The minimal gains, even hurting \TER when 23k samples are used.
It could be attributed to the fact that 23k segments are not sufficient to train a reliable deep model. 
% Using 23K samples, however, the improvement was much less than the other tasks, even it hurts \TER score.
We believe that as long as a moderate amount of data is available, SpecAugment helps data efficiency more.
The augmentation policy compensates lack of data when we half the training data size to 47k segments.
It achieves 1.8\% of absolute improvement difference in \BLEU and 2.8\% in \TER
compared to the model trained using 94k samples with 1.5\% of \BLEU and 1.6\% of \TER.

\subsection{Pre-training}

We also consider the effect of data augmentation on the top of pre-training.
We pre-train the encoder using our pre-trained ASR model,
and the decoder using our MT model as described in Section \ref{sec:models} which use more training data compared to the direct ST model (see Table \ref{tab:stat}).
After initialization with pre-trained components, the ST model is fine-tuned using the ST data.
Here, we add an additional BLSTM layer (adaptor layer) to adapt the output of speech encoder
and the input of text decoder without freezing the parameters (see Fig. \ref{fig:direct}). 

As shown in Table \ref{tab:specAug_pretraining}, SpecAugment slightly outperforms the pre-training.
It outperforms the pre-trained models by 0.5\% and 0.2\% \BLEU on LibriSpeech and IWSLT respectively,
and no \TER improvements.
This results confirm that SpecAugment can be used along with pre-training.
By comparing Table \ref{tab:specAug_libri} and \ref{tab:specAug_pretraining},
one can argue that the SpecAugment might compensate the effect of pre-training strategy
by its own for LibriSpeech
(compare 18.0\% vs. 17.9\% in \BLEU on the dev set and 15.8\% vs. 16.1\% on the test set).

We finally compare our model with the other works in the literature in Table \ref{tab:others}.
On the LibriSpeech test set,
our model outperforms both the LSTM-based end-to-end models
and the Transformer-based. 
Contrary to \cite{berard_2018_librispeech} in which character decoder is used,
we apply BPE that obtain improvements.
Both our direct model and the cascade model outperform the models in \cite{berard_2018_librispeech}.
We also beat the Transformer models without augmentation.
Our recipe works as good as knowledge distillation method where an MT model is exploited to teach the ST model.

\begin{table}
\begin{center}
\caption{Comparison on LibriSpeech En$\to$Fr test set with the literature.
In order to be comparable with other works,
the results in this table are case-insensitive \BLEU computed using \texttt{multi-bleu.pl} script \cite{koehn_07_moses}.
$^1$: the evaluation is without punctuation.
$^2$: it correspond to 16.2\% \BLEU from Table \ref{tab:specAug_pretraining}.}
\label{tab:others}
\begin{tabular}{lcc}

\toprule
Method & \BLEU[\%] \\ 
\midrule
\bfseries other works  \\ 
\quad direct \cite{berard_2018_librispeech}         & 13.3\\
\quad multi-task  \cite{berard_2018_librispeech}     & 13.4\\
\quad cascade pipeline  \cite{berard_2018_librispeech}  & 14.6\\
\quad unsupervised$^1$   \cite{chung_2019_unsupervised_st}   & 12.2  \\ 
\quad Transformer \cite{gangi-etal-2019-enhancing}   & 13.8\\
\quad Transformer+pretraining   \cite{st_kd_interspeech2019}   & 14.3 \\
\quad\quad + knowledge distillation   \cite{st_kd_interspeech2019}   & 17.0  \\

\toprule
\bfseries this work  \\  
% \quad cascade pipeline  & 24.43  & 15.74\\
\quad direct+pretraining+SpecAugment$^2$   & 17.0 \\ 

\bottomrule
\end{tabular}
\end{center}
\end{table}

\section{Conclusion} \label{conclude}

We have studied SpecAugment,
a simple and low-cost data augmentation for end-to-end direct speech translation.
There is a limit of how much data augmentation can be applied.
Adding some data augmentation improves the performance in terms of \BLEU and \TER,
however, at some point, the performance degrades by more augmentation. 
We have also shown that the method avoids overfitting to some extent and it requires longer training.
A common criticism of many techniques for low-resource applications is
that the improvements go away once we have lots of synthetic parallel data.
Therefore, we believe that comparing the SpecAugment approach
with generated data using TTS model is a crucial next step.
We also aim to explore the effectiveness of the approach on Transformer architecture. 

\section{Acknowledgements}
\label{sec:acknowledgements}
\begin{wrapfigure}{l}{0.08\textwidth}
\vspace{-5mm}
    \begin{center}
      \includegraphics[width=0.12\textwidth]{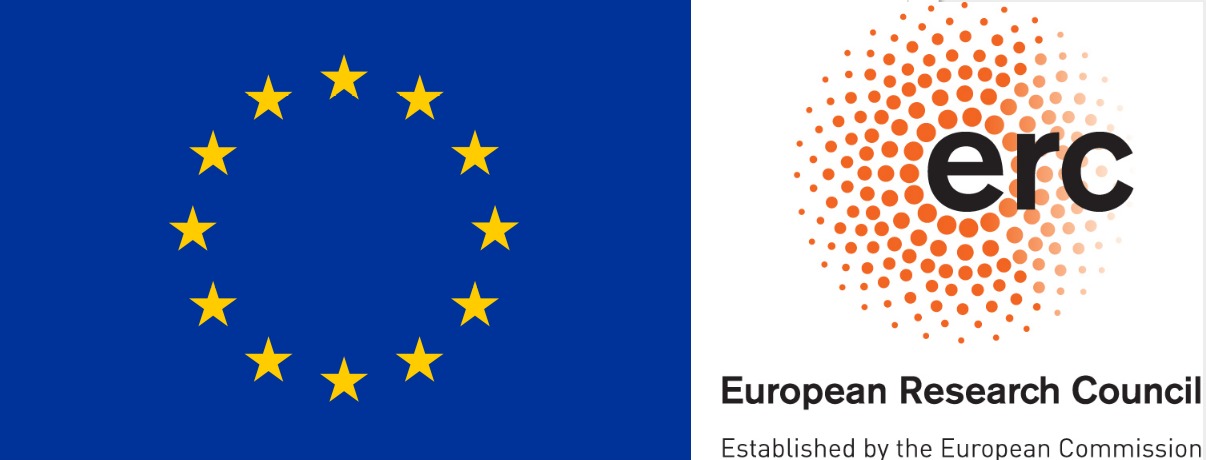} \\
      \vspace{2mm}
      \includegraphics[width=0.12\textwidth]{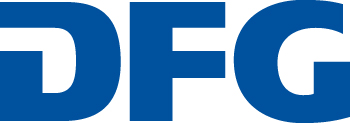}
    \end{center}
\vspace{-4mm}
\end{wrapfigure}
This work has received funding from the European Research Council (ERC) under the European Union's Horizon 2020 research and innovation programme (grant agreement No 694537, project "SEQCLAS"), the Deutsche Forschungsgemeinschaft (DFG; grant agreement NE 572/8-1, project "CoreTec") and from a Google Focused Award. The work reflects only the authors' views and none of the funding parties is responsible for any use that may be made of the information it contains. 

\bibliographystyle{IEEEtran}

% Give us some more space:
% Tune this later if needed, or just uncomment if not needed.

%\linespread{0.5}
%\linepenalty=10000

% This works perfect! :)
%\setstretch{0.85}

%\def\baselinestretch{0.8}
%\let\normalsize\footnotesize\normalsize
%\SetTracking{encoding=*}{-15}\lsstyle  % still somewhat ok
%\SetTracking{encoding=*}{-85}\lsstyle

%\renewcommand{\baselinestretch}{0.1}\normalsize
% http://tex.stackexchange.com/questions/93859/condense-the-space-between-bibliographic-entries
\let\OLDthebibliography\thebibliography
\renewcommand\thebibliography[1]{
  \OLDthebibliography{#1}
  \setlength{\parskip}{0pt}
  \setlength{\itemsep}{0pt plus 0.07ex}
}

\bibliography{iwslt2018-submission}

%
% \bibliographystyle{IEEEtran}
% \begin{thebibliography}{10}
% \bibitem[1]{ES1} Smith, J. O. and Abel, J. S.,
% ``Bark and {ERB} Bilinear Transforms'',
% IEEE Trans. Speech and Audio Proc., 7(6):697--708, 1999.
% \bibitem[2]{ES2} Lee, K.-F., Automatic Speech Recognition:
% The Development of the
% SPHINX SYSTEM, Kluwer Academic Publishers, Boston, 1989.
% \bibitem[3]{ES3} Rudnicky, A. I., Polifroni, Thayer, E. H.,
%  and Brennan, R. A.
% "Interactive problem solving with speech", J. Acoust. Soc. Amer.,
% Vol. 84, 1988, p S213(A).
% \end{thebibliography}
\end{document}